\documentclass[runningheads]{llncs}

\usepackage{amsmath,epsfig,amssymb, amsbsy,bm}
\usepackage{graphicx,centernot}
\usepackage{subcaption}

\usepackage{floatrow}
\usepackage{blindtext}

\newfloatcommand{capbtabbox}{table}[][\FBwidth]

\newcommand{\CI}{\mathrel{\perp\mspace{-10mu}\perp}}

\usepackage{xcolor}
\usepackage{soul}
 
\begin{document}
\pagestyle{empty} 

\mainmatter

\title{The color of smiling: computational synaesthesia of facial expressions}
\titlerunning{Lecture Notes in Computer Science}

\author{Vittorio Cuculo \inst{1} \and Raffaella Lanzarotti \inst{2} \and Giuseppe Boccignone \inst{2}} 

\authorrunning{Vittorio Cuculo}

\institute{Dipartimento di Matematica
 Universit\`a di Milano\\
via Cesare Saldini 50, Italy\\
\email{vittorio.cuculo@unimi.it}\\\and Dipartimento di Informatica
 Universit\`a di Milano\\
via Comelico 39/41, Italy\\
\email{\{lanzarotti,boccignone\}@di.unimi.it}\\
}

\maketitle
\begin{abstract}
This note gives a preliminary account of the transcoding or rechanneling problem between different stimuli as it is of interest for the natural interaction or affective computing fields.
By the consideration of a simple example, namely the color response of an affective lamp to a sensed facial expression, we frame the problem within an information-theoretic perspective. 
A full justification in terms of the Information Bottleneck principle promotes a latent affective space, hitherto surmised as an appealing and intuitive solution, as a suitable mediator between the different stimuli.
\end{abstract}

\section{Introduction}
At the heart of non-verbal interaction between agents, either artificial or biological, is a rechannelling ability, namely the  ability
of gathering data from one kind of signal and instantaneously turn it into a different kind of signal. In artificial agents, such rechannelling or transcoding ability must be simulated through some form of ``computational synaesthesia''. Strictly speaking,  synaesthesia  is a neurological phenomenon in which stimulation of one sensory or cognitive pathway leads to automatic, involuntary experiences in a second sensory or cognitive pathway \cite{ward2004synaesthesia}. Here, more liberally, we adopt it as a good metaphor for such rechannelling/transcoding of information \cite{scheirer2000affective}. 
 
In this note, as a case study, we consider the problem of transducing a sensed facial expression into a color stimuli. Denote $\mathbf{V}$ and $\mathbf{C}$ the random variables (RVs) standing for a visible expression display and for an emitted  color stimulus, respectively. Then, the transcoding $\mathbf{V} \mapsto \mathbf{C}$ can be described in probabilistic terms as that of sampling  a specific color stimulus  $\mathbf{c}$, when expression stimulus $\mathbf{v}$ is observed, namely 
\begin{equation}
\mathbf{c} \sim P(\mathbf{C} \mid \mathbf{V}= \mathbf{v}),
\label{eq:trans}
\end{equation}
\noindent where $P(\mathbf{C} \mid \mathbf{V})$ is the conditional probability density function (pdf) defining the probability of generating a color stimulus $\mathbf{c}$  conditioned on  the observation of expression  $\mathbf{v}$. Such kind of problem  is of interest for many applications in social signal processing \cite{Vinciarelli2013}, natural interaction \cite{scheirer2000affective}, social robotics \cite{Kim2008}. But, most important, here we discuss how a principled solution involves deep issues  in spite of the apparent specificity of the problem.  

An appealing  way to conceive transcoding is through the mediation of some kind of latent space  in particular a space of affective or emotional experience,  which confers a unified semantics to the different kinds of non-verbal signals. It has been argued  that this could be necessary for grounding synaesthetic cross-modal correspondences \cite{spence2011crossmodal,collier1996affective}, simulation-based theory of emotion and empathy \cite{vitale2014affective}. Also, the mediation of a continuous dimensional space has been advocated for analyzing many different expressive modalities and to the purpose of building affective objects \cite{scheirer2000affective}. To such aim, we focus on the Pleasure/Arousal/Dominance space (PAD, \cite{russell1977evidence}) as a continuous  latent  space to support ``synesthesia'' of facial expressions into color.

In this study we discuss how such solution can be conceived and grounded in an information-theoretic perspective, namely the Information Bottleneck (IB) framework introduced in \cite{tishbyinformation} (cfr. Section \ref{sec:back}).




\begin{figure}[htb]
\centering
\includegraphics[scale=0.22]{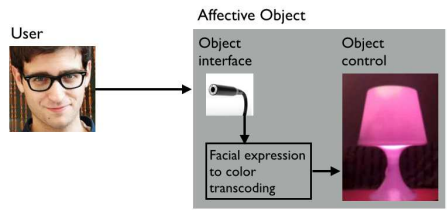}
\caption{The Mood Lamp: an affective ``synaesthetic'' object that responds to user's  facial  expressions by changing the color of the light emitted.
} 
\label{Fig:MoodL}
\end{figure}

As a proof of concept we present the Mood Lamp (cfr. Fig. \ref{Fig:MoodL}). The Mood Lamp is a kind of \emph{affective object}, that is a ``physical object which has the ability to sense emotional data from a person, map that information to an abstract form of expression and communicate that information expressively, either back to the subject herself or to another person''\cite{scheirer2000affective}. In particular, here a facial expression is used to convey  affect states to an Ikea RGB color lamp, which  will respond  by changing the color of the light emitted in accordance with the  affect. 



Modeling computational synaesthesia as specified through Eq. \ref{eq:trans} in the IB perspective has the advantage of providing a principled approach, characterized by a minimum of assumptions (Section \ref{sec:method}). However there are a number of subtle difficulties to overcome that deserve being discussed (cfr. Sections \ref{sec:facemood} and \ref{sec:moodcolor}).

\section{Background and rationales}
\label{sec:back}



Central to this work is the idea that the synaesthetic transduction 
$\mathbf{V} \mapsto \mathbf{C}$ can be performed  by resorting to an affect space, say $\mathbf{E}$,  as a mediating factor.

Resorting to affect for transcoding stimuli may seem \emph{prima facie}  an instrumental approach;  however, two issues bear on this choice. First, the insight of an affect space as a common factor for rechanneling between kinds of information is a not new in the psychological literature. On the one hand, perception and emotion are closely linked \cite{pessoa2008relationship}.
For instance, as to the specific case of  synaesthetic cross-modal correspondences, affective similarity \cite{spence2011crossmodal} has  been suggested as a contributing factor:  stimuli may be matched  if they both happen to increase an observer's level of alertness or arousal, or if they both happen to have the same effect on an observer's emotional state, mood, or affective state. Efficient handling of affective synesthesia has been discussed by Collier who has shown  \cite{collier1996affective} that both  perceptual stimuli -- such as colours, shapes, and musical fragments -- and human emotions can be represented in a simple multidimensional space with two or three corresponding dimensions. Clearly this idea is consistent with the framework of an underpinning continuous ``affect space'', which can be approximated by  either two primary dimensions, e.g. valence and activity (arousal) \cite{osgood1964measurement},  or three such as pleasure, arousal, and dominance (PAD)  as proposed by  Russell and Mehrabian \cite{russell1977evidence} .

\begin{figure}[ht]
\begin{subfigure}{.5\textwidth}
  \centering
  \includegraphics[width=.99\linewidth]{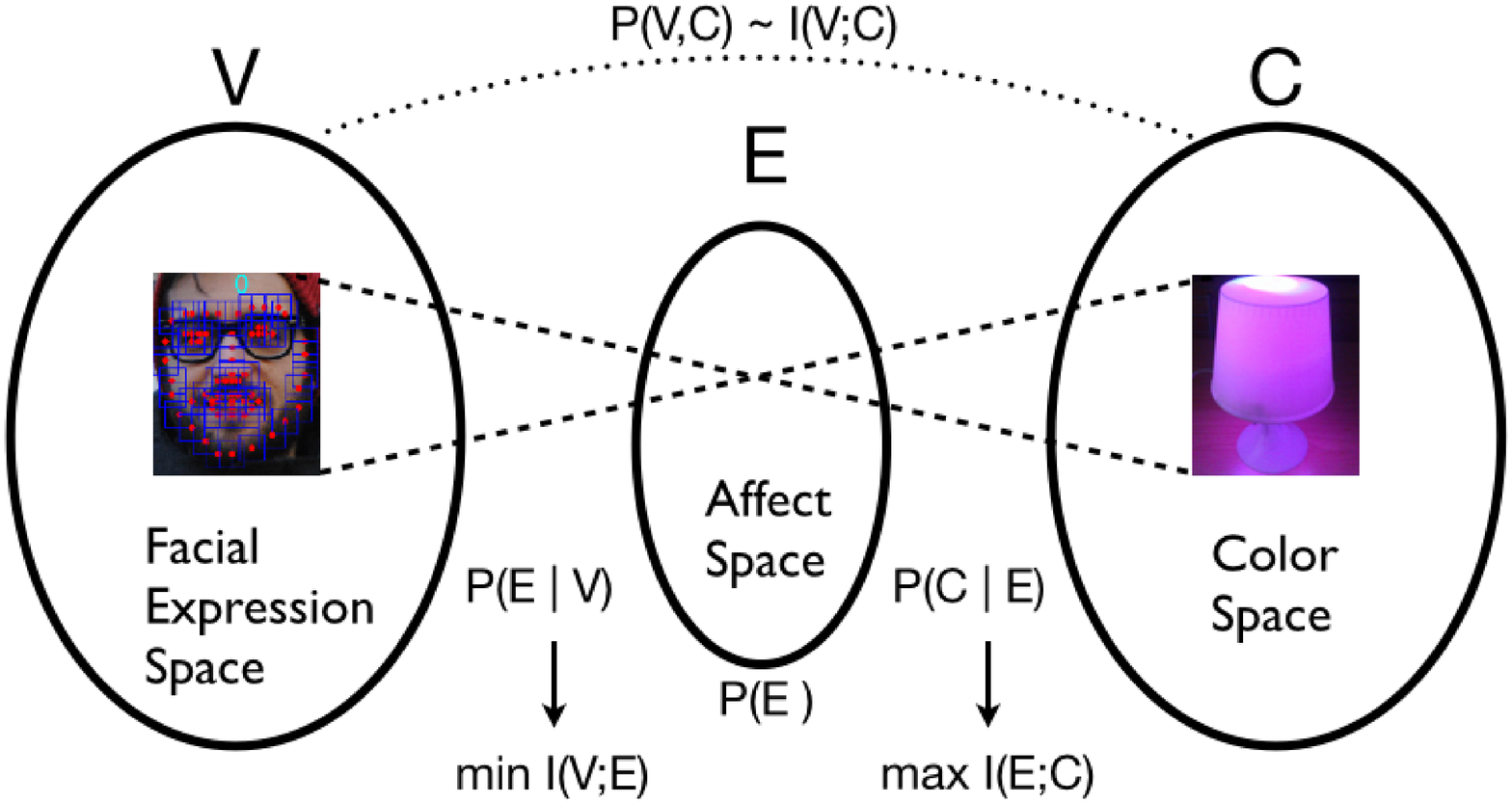}
  \caption{\textbf{(a)} The IB framework}
  \label{Fig:FigIB}
\end{subfigure}%
\begin{subfigure}{.5\textwidth}
  \centering
  \includegraphics[width=.99\linewidth]{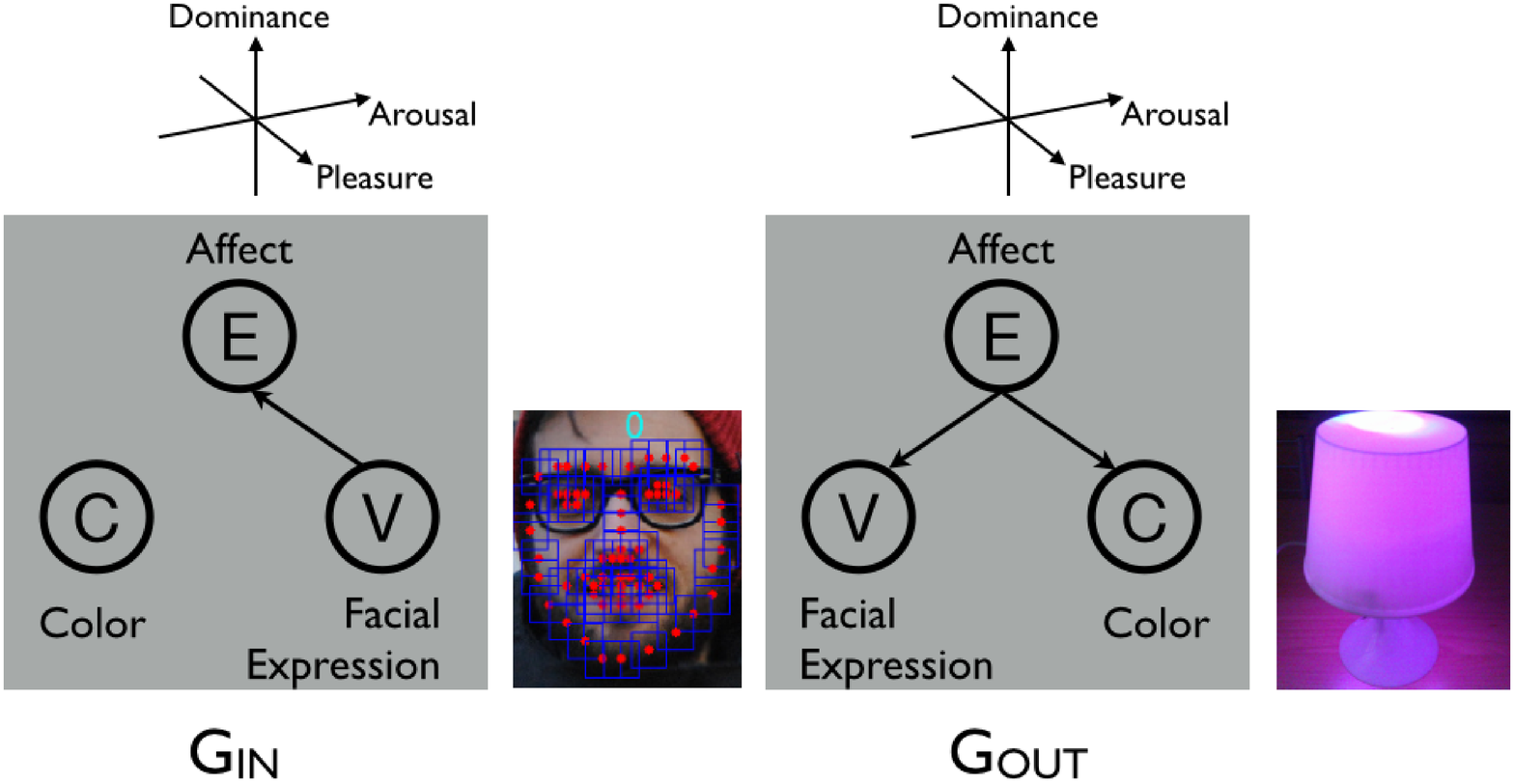}
  \caption{\textbf{(b)} A PGM representation of IB}
  \label{Fig:FigGM}
\end{subfigure}
\caption{Synesthesia of facial expression $\mathbf{V} $ into the (lamp) color  $\mathbf{C}$   as an Information Bottleneck problem. The displayed expression is represented as a random vector $\mathbf{V} $, computed on the basis of facial landmarks $\mathbf{L}$ (displayed as red dots superimposed on the face).
\textbf{(a)} Transcoding $\mathbf{V} \mapsto \mathbf{C}$ is modelled as the search for a compressed representation   of $\mathbf{V}$, namely the affect space $\mathbf{E}$, which achieves minimum redundancy while maintaining the mutual information $I(\mathbf{E}; \mathbf{C})$ about the relevant variable $\mathbf{C}$, as high as possible.  
\textbf{(b)} The left graph $\mathcal{G}_{IN}$ encodes the compression process; the right graph $\mathcal{G}_{OUT}$ is the target model representing which relations should be maintained or predicted. The IB principle boils down to minimize the information maintained by $\mathcal{G}_{IN}$ and to maximize the information preserved by $\mathcal{G}_{OUT}$. 
}
\label{Fig:FigIBFull}
\end{figure}

The second issue, which is tied in a subtle way to the previous one, grounds in the general  and fundamental principle that an organism who maximizes the  adaptive value of its actions given fixed resources should have internal representations of the outside world that are optimal in a very specific information-theoretic sense \cite{bialek2006efficient}. In a communicative action,  this optimization problem is related to joint source channel coding, namely the task of encoding and transmitting information simultaneously in an efficient manner \cite{cover}.

One route to do  justice to both issues is the  Information Bottleneck (IB), \cite{tishbyinformation}. IB is an information-theoretic principle for coping with the extraction of relevant components of an ``input'' random variable $\mathbf{X}$, with respect to an ``output'' random variable $\mathbf{Y}$. This is performed by finding a \emph{bottleneck} variable, that is a compressed, non-parametric and model-independent representation $\mathbf{T}$ of $\mathbf{X}$, that is most informative about $\mathbf{Y}$.

In our case the intuition is that the  bottleneck  variable $\mathbf{E}$ is suitable to capture the relevant  affective aspects of the facial expression stimuli $\mathbf{V}$ that are informative about the output color stimulus $\mathbf{C}$ (cfr. Fig. \ref{Fig:FigIB}). 

Denote $I(\mathbf{X};\mathbf{Y})$ 
the   mutual information  \cite{cover}. 
The original IB approach determines the auxiliary latent space $\mathbf{E}$ and related mapping $\mathbf{V}\mapsto \mathbf{E}$, such that  the mutual information  $I(\mathbf{V};\mathbf{E})$ is minimized (to achieve maximum compression),  while relevant information $I(\mathbf{E};\mathbf{C})$ is maximized. Hence  
\begin{equation}
\min_{\mathbf{V}\mapsto \mathbf{E}} I(\mathbf{V};\mathbf{E}) -\beta I(\mathbf{E};\mathbf{C}),
\label{eq:IB}
\end{equation}
\noindent where $\mathbf{V}\mapsto \mathbf{E}$ is the rule for creating the internal representation, and the positive parameter $\beta$ smoothly controls the tradeoff between compression and preserved relevant information.


The optimization principle in Eq. \ref{eq:IB} is very abstract; also, no analytical solution is available. However,  it has been shown by Friedman et al. \cite{friedman2001multivariate} that the IB problem can be suitably reformulated in terms of directed Probabilistic Graphical Model (PGM, \cite{murphy2012machine}) representation (cfr. Fig. \ref{Fig:FigGM}).
A directed PGM  is a graph-based representation  where nodes denote RVs and  arrows/arcs code conditional dependencies between RVs. 
Stated technically, the $\mathcal{G}$ structure encodes  the set 
of conditional independence assumptions over  the set of RVs $\{\mathbf{X}_i\}$ (called the local independencies, \cite{murphy2012machine}) involved by the joint pdf $P(\{\mathbf{X}_i\})$ associated to $\mathcal{G}$.
Then,  the  joint pdf factorizes according to $\mathcal{G}$ \cite{murphy2012machine}, that is $P$ is consistent with $\mathcal{G}$, $P \models \mathcal{G}$. Given a PGM $\mathcal{G}$,   $I^{\mathcal{G}} = \sum_i I(\mathbf{X}_i; Pa_{i}^{\mathcal{G}})$ denotes the information  computed with respect to the pdf $P \models \mathcal{G}$ \cite{friedman2001multivariate}, where $Pa_{i}^{\mathcal{G}}$ stands for the ensemble of  parents of node $\mathbf{X}_i$.

Under these circumstances, the IB principle (Eq. \ref{eq:IB}) can be shaped in the language of PGMs by considering two directed graphs $\mathcal{G}_{IN}$ and $\mathcal{G}_{OUT}$, together with the pdfs entailing such graphs, $Q \models \mathcal{G}_{IN}$ and $P \models \mathcal{G}_{OUT}$, respectively (cfr. Fig. \ref{Fig:FigGM}).  
Thus, the information that we would like to minimize is now given by $I^{\mathcal{G}_{IN}}$, where $I^{\mathcal{G}_{IN}} = I(\mathbf{V},\mathbf{E})$.
The relevant information that we wish to preserve is specified by  the \emph{target model} $\mathcal{G}_{OUT}$, as $I^{\mathcal{G}_{OUT}} = I(\mathbf{E}; \mathbf{V})+ I(\mathbf{E};\mathbf{C})$. Assuming this, Eq. \ref{eq:IB} can be rewritten,
\begin{equation}
\min_{\mathbf{V}\mapsto \mathbf{E}} I^{\mathcal{G}_{IN}} -\beta I^{\mathcal{G}_{OUT}}=
\min_{Q} I(\mathbf{V};\mathbf{E}) + \gamma D_{KL}(Q(\mathbf{V}, \mathbf{E},\mathbf{C}) || P(\mathbf{V}, \mathbf{E},\mathbf{C}))
\label{eq:IB2}
\end{equation}
\noindent where  $D_{KL}(Q(\mathbf{X}) || P(\mathbf{X}))$ 
is the Kullback-Leibler divergence between distributions $Q$ and $P$  \cite{cover}. 
The scale parameter $\gamma$ balances the above two factors and is related to $\beta$ as $\beta = \gamma / (1+\gamma)$. In the limit $\gamma \to 0$  we are only interested in compressing the variable  $\mathbf{V}$. When $\gamma \to \infty$  we concentrate on choosing a pdf $Q$ that is close to the distribution $P \models \mathcal{G}_{OUT}$:
\begin{equation}
P(\mathbf{V},\mathbf{C},\mathbf{E})= P(\mathbf{V} \mid \mathbf{E}) P(\mathbf{C} \mid \mathbf{E}) P(\mathbf{E}),
\label{eq:Pjoint}
\end{equation}
\noindent by minimizing $D_{KL}(Q(\mathbf{V}, \mathbf{E},\mathbf{C}) || P(\mathbf{V}, \mathbf{E},\mathbf{C}))$.



It has been shown that iterative approximate solutions to Eq. \ref{eq:IB}, which cycle between determining $Q(\mathbf{E})$  and $Q(\mathbf{C} \mid \mathbf{E})$ for a fixed $Q(\mathbf{E} \mid \mathbf{V})$, and computing $Q(\mathbf{E} \mid \mathbf{V})$ for fixed $Q(\mathbf{E})$  and $Q(\mathbf{C} \mid \mathbf{E})$, are a formulation  of the generalized Expectation-Maximization algorithm for clustering \cite{slonim2002maximum}. Clearly, this holds when the latent space $\mathbf{E}$ is a discrete space. Indeed, it is readily seen that at the extreme spectrum $\gamma \to \infty$, the minimization in Eq. \ref{eq:IB2} boils down to minimize $D_{KL}(Q(\mathbf{V}, \mathbf{E},\mathbf{C}) || P(\mathbf{V}, \mathbf{E},\mathbf{C}))$ which is but one instance of the Variational Bayes method for learning the generative model $P \models \mathcal{G}_{OUT}$, as represented in the target model of Fig. \ref{Fig:FigGM}.

When the transcoding operation relies upon a continuous latent space -- as in our case -- the IB approach represented in terms of $P \models \mathcal{G}_{OUT}$ is reminiscent of several latent factor models for paired data, such as Bayesian factor regression, Probabilistic Partial Least squares and Probabilistic CCA \cite{murphy2012machine}. 

\section{Methods}
\label{sec:method}
The IB approach provides a principled justification to the use of  a mediating latent space for simulating computational synaestesia. After the learning stage, when the  distribution factors of the target joint pdf are available, transcoding in Eq. \ref{eq:trans} can be performed via the latent space  $\mathbf{E}$:

\begin{equation}
\mathbf{e} \sim P(\mathbf{E} \mid \mathbf{V}= \mathbf{v}), \quad
\mathbf{c} \sim P(\mathbf{C} \mid  \mathbf{E}= \mathbf{e}).
\label{eq:trans2}
\end{equation}
It is worth remarking that learning procedures implementing   optimization  (\ref{eq:IB}) or (\ref{eq:IB2}) have the goal of designing from scratch a latent space that is optimal with respect to the given constraints and the  joint  distribution, here $P(\mathbf{V},  \mathbf{C})$. In the case study we are considering, conditions are slightly different. First, the latent space $\mathbf{E}$ is not constructed abstractly, but it should be chosen guided by psychological theories of emotion; this somehow simplifies some machine learning issues, for instance, the dimensionality of the space is not to be learned. Second, the joint pdf is not straightforwardly available.

As to the first issue, we assume a core affect representation. Core affect is a neurophysiological state that underlies simply feeling good or bad, drowsy or energised, and it can be experienced as free-floating, or mood, or can be attributed to some cause (and thereby begin an emotional episode) \cite{russell2003core}. Thus, it is a continuous latent space and a suitable representation is provided by the PAD  space proposed by Mehrabian and Russell \cite{russell1977evidence}. Such space can be described along three nearly independent continuous dimensions: Pleasure-Displeasure (measured by $P$), Arousal-Nonarousal ($A$), and Dominance-Submissiveness ($D$); thus, $\mathbf{E} = \left[ P A D \right]^{\intercal}$. 

Note that, under the assumption of an actual affective state $\mathbf{E}=\mathbf{e}$, it is easy to show, by using Bayes' rule and the joint pdf factorisation (\ref{eq:Pjoint}), that $P(\mathbf{V},\mathbf{C} \mid \mathbf{E})= P(\mathbf{V} \mid \mathbf{E}) P(\mathbf{C} \mid \mathbf{E})$, thus $\mathbf{V} \CI \mathbf{C} \mid \mathbf{E}$. That is, if the affective state is given, then $\mathbf{V}$  and $\mathbf{C}$ are conditionally independent.
The very issue here is thus obtaining the ``mapping'' probabilities $P(\mathbf{E}\mid \mathbf{V})$ and $P(\mathbf{C} \mid \mathbf{E})$. 
To this end, we can make the simplifying assumption of a Gaussian IB \cite{chechik2005information}. In this case an optimal compression $\mathbf{E}$ is obtained with a noisy linear transformation of $\mathbf{V}$:
\begin{equation}
\mathbf{e} = \mathbf{W}_{E} \mathbf{v} + \xi_{E}, \quad \quad \xi_{E} \sim \mathcal{N}(\mathbf{0},\Sigma_{\xi_{E}}),
\label{eq:padreg}
\end{equation}
\noindent where $\xi_{E}$ is an additive noise term sampled from a zero-mean Gaussian pdf  $\mathcal{N}(\mathbf{0},\Sigma_{\xi_{E}})$. 

Similarly,  the most natural choice for color is a continuous  space; e.g., in studies concerning relationships between color and emotion the HSL space -- defined on Hue ($H$),  Saturation ($S$) and Luminance ($L$) -- has been used \cite{Gao2006,Kim2008}. Thus, a generative model for mapping  $P(\mathbf{V} \mid \mathbf{E})$ is 
\begin{equation}
\mathbf{c} = \mathbf{W}_{C} \mathbf{e} + \xi_{C}, \quad \quad \xi_{C} \sim \mathcal{N}(\mathbf{0},\Sigma_{\xi_{C}}).
\label{eq:hslreg}
\end{equation}

Eqs. \ref{eq:padreg} and \ref{eq:hslreg} nicely simplify the synaesthetic mapping to a pair of regressions on a joint latent space, however the second issue related to the actual availability of $P(\mathbf{V}, \mathbf{C})$ must be taken into account.
Needless to say, the use of a  continuous affect space brings along a number of challenges. In the psychological literature, fleeting changes in the countenance of a face are considered to be  ``expressions of emotion'' (EEs) 
and have been systematically investigated by Ekman  \cite{ekman1993facial} in a categorical perpsective. Ekman's  work has  fostered  a vaste amount of theoretical and empirical work, which has been particularly influent in the affective computing community \cite{Vinciarelli2013}. Under these circumstances, finding the map $\mathbf{V} \mapsto \mathbf{E}$, has been mostly relied on a pattern recognition approach to infer emotions from expressions  under the fundamental assumption of basic emotions, for example by considering the discrete set $\mathbf{E}= \{ \text{joy},\text{sadness}, \text{anger},\text{disgust}, \text{surprise},\text{fear}\}$. By contrast, Eq. \ref{eq:padreg} assumes a probabilistic relationship between $\mathbf{E}$ and $\mathbf{V}$ where $\mathbf{E}$ is continuously defined.

A second problem to solve is related to Eq. \ref{eq:hslreg}, that is to learn the mapping $\mathbf{E} \mapsto \mathbf{C}$.  In the past decades, only a few researchers investigated the relationship between color and emotion \cite{mehrabian1994,Mahnke1996,Birren2006,Kim2008,Suk2010} (and often in the sense of emotion elicited by a colors and not the vice versa). In this case, the main problem is setting up a minimal training set which we derive from data available from the psychological literature. These issues are addressed in the following sections.

\section{From face expression to mood}
\label{sec:facemood}

In this section we detail how we solve the problem of learning a probabilistic relationship between $\mathbf{E}$ and $\mathbf{V}$ where $\mathbf{E}$ is continuously defined according to the PAD  model. To this end, we exploit results of experimental studies that have evaluated the PAD value of  discrete emotion states, e.g.,  \cite{hoffmann2012mapping}. 


A very first step concerns with the facial landmark localisation, which can
be summarised as follows. Denote $\mathbf{L}= \{\mathbf{l}^1,
\mathbf{l}^2,\cdots, \mathbf{l}^n\}$  the locations of $n$
landmarking parts of the face, and $\mathbf{F}= \{\mathbf{f}^1,
\mathbf{f}^2,\cdots, \mathbf{f}^n\}$ the measured detector
responses, where $\mathbf{f}^i = \phi(\mathbf{l}^i, \mathcal{I})$
is the response or feature vector provided by a local detector at
location $\mathbf{l}^i$ in image or frame $\mathcal{I}$. Then,
localisation can be solved by finding the value of $\mathbf{L}$
that maximises the probability of $\mathbf{L}$ given the responses
from local detectors, namely $\mathbf{L}^{*}= \arg
\max_{\mathbf{L}} P(\mathbf{L} | \mathbf{F})$. Following \cite{CucLaBo_Euvip11},  we  exploit a part-based framework that integrates  an effective local representation  based on   sparse coding.  Sparse coding has recently gained currency in face analysis (e.g.,\cite{Jeni2013,GrossiLanzaottiICIP2013}).  
In particular:
\begin{equation}
\mathbf{L}^{*}= \arg \max_L \sum_{k=1}^{m} \int_{t} \prod_{i=1}^{n} P(\Delta \mathbf{l}^i_{k,t})  P(\mathbf{l}^i |\mathbf{f}^i) dt,
\label{eq:L}
\end{equation}
\noindent where the prior $P(\Delta \mathbf{l}^i_{k,t})$ accounts for the \emph{shape} or global component of the model, and $P(\mathbf{l}^i |\mathbf{f}^i))$ for the \emph{appearance} or local component. 
For what concerns the local component $P(\mathbf{l}^i|\mathbf{f}^i)$,  
 we resort to  Histograms of Sparse Codes to sample  patch responses $\mathbf{f}^i$,
which we learn from facial images (see \cite{CucLaBo_Euvip11} for details). 







\begin{figure}
\begin{floatrow}
\ffigbox{%
  \includegraphics[scale=0.18]{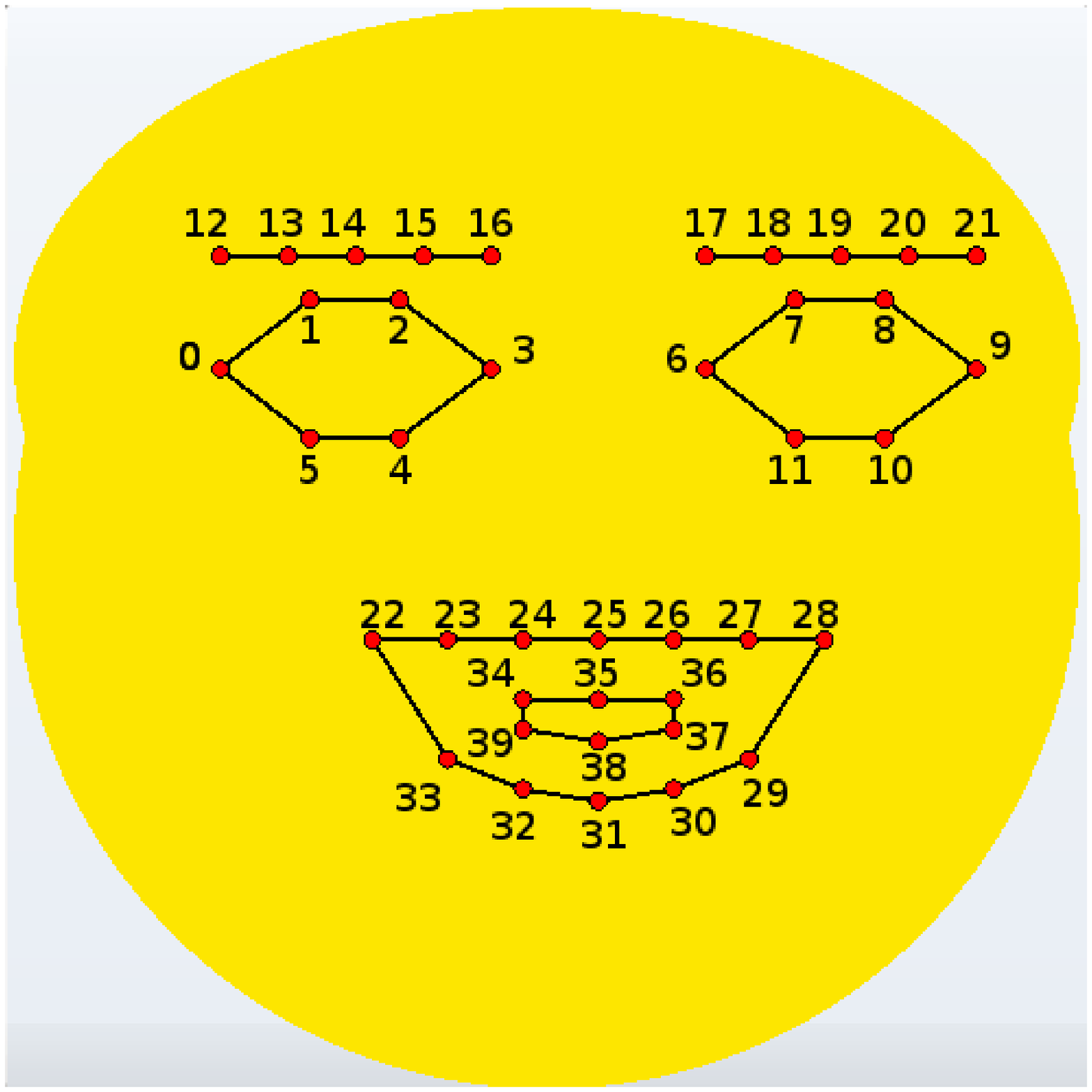}
}{%
  \caption{The 40 facial landmarks}%
  \label{fig:land}
}
\capbtabbox{%
\scriptsize
\begin{tabular}{|c|c|c|}
\hline
\textbf{Name} & \textbf{EP} & \textbf{Definition} \\
\hline
Eyes height & $\mathbf{v}^{0}$ & $(\mathbf{l}^{0}_{y} - \mathbf{l}^{1}_{y}) * 2$ \\
\hline
Eyes / brows space   & $\mathbf{v}^{1}$ & $\mathbf{l}^{0}_{y} - \mathbf{l}^{13}_{y} + \frac{(\mathbf{l}^{16}_{y} - \mathbf{l}^{12}_{y})}{4}$ \\
\hline
Eyebrow's inner height & $\mathbf{v}^{2}$ & $\mathbf{l}^{12}_{y} + \mathbf{v}^{1} - \mathbf{l}^{0}_{y}$ \\
\hline
Eyebrow's outer height & $\mathbf{v}^{3}$ & $\mathbf{l}^{16}_{y} + \mathbf{v}^{1} - \mathbf{l}^{0}_{y}$ \\
\hline
Mouth width & $\mathbf{v}^{4}$ & $\mathbf{l}^{28}_{x} - \mathbf{l}^{22}_{x}$· \\
\hline
Mouth openness & $\mathbf{v}^{5}$ & $\mathbf{l}^{31}_{y} - \mathbf{l}^{25}_{y}$ \\
\hline
Mouth twist & $\mathbf{v}^{6}$ & $\frac{\mathbf{l}^{28}_{y} - \mathbf{l}^{25}_{y}}{2} - \mathbf{l}^{22}_{y}$ \\
\hline
\end{tabular}
}{%
  \caption{Visual expression parameters (EP) via local landmark displacements}%
  \label{tab:fpar}
}
\end{floatrow}
\end{figure}

For each image/frame we consider $40$ landmarks $\mathbf{L}= \left[ \mathbf{l}^{1} \cdots \mathbf{l}^{40} \right]^{\intercal}$ as shown in Fig. \ref{fig:land}, and we map them into a vector of visible expression parameters $\mathbf{V}$ by measuring the landmark displacements. This step, in a vein similar to Action Units approaches \cite{ekman1993facial}, is aimed at capturing the expression movements within local face region, such as mouth-bent, eye-open and eyebrow-raise, etc., as detailed in Tab. \ref{tab:fpar} \cite{Broekens2013}.

The extracted expression parameters $\mathbf{V}\in\mathbb{R}^{7}$ are put in correspondence to PAD values, $\mathbf{E}\in\mathbb{R}^{3}$, by using Eq. \ref{eq:padreg}. In the current simulation,  a multilinear ridge regression has been used, that is a penalized least squares method that adds a Gaussian prior to the parameters  to encourage them to be small. Such model has interesting connection to latent variable space inference \cite{murphy2012machine}. 



\section{The color of mood}
\label{sec:moodcolor}


Here, we discuss some subtleties related to  the mapping $\mathbf{E} \mapsto \mathbf{C}$ in order to  learn the generative model of Eq. \ref{eq:hslreg}. Recall that we represent color as a random vector in HSL color space, i.e. $\mathbf{c}= \left[ H S L \right]^{\intercal}$.


%
%
%

The seminal work investigating the relationship between color and emotion is that by Valdez and Mehrabian \cite{mehrabian1994}. They mainly studied how saturation $S$ and luminance $L$  affect PAD.  In \cite{Mahnke1996,Birren2006,Suk2010} the emotions elicited by basic colors have been qualitative presented. 
Only recently in \cite{Kim2008} a synthesis of these approaches has been proposed, aiming at allowing robots to express the intensity of emotions by coloring and blinking LED placed around their eyes.
This work has the limit to resort to only two distinct values for both $S$ and $L$, and four values for the hue $H$, hence leading different emotions to be represented by the same color.

In our study, we propose a finer correspondence model, preserving  maximum representativeness of the three components HSL.
More precisely, as to $S$ and $L$, following \cite{Kim2008}, we invert the dependency of PAD values proposed in  \cite{mehrabian1994}, while maintaining the obtained results. Define
\begin{equation}
P = 0.69  L + 0.22  S , \quad
A = -0.31  L + 0.60  S , \quad
D = -0.76  L + 0.32  S.
\end{equation}
\noindent Then, $S$ and $L$ can be derived
\begin{equation}
\hat{\mathbf{c}} = (\mathbf{W}^{\intercal} \mathbf{W})^{-1} \mathbf{W}^{\intercal} \mathbf{e},
\label{eq:part}
\end{equation}
\noindent where $\hat{\mathbf{c}}=\begin{bmatrix}L S\end{bmatrix}^{\intercal}$,   $\mathbf{W} = \begin{bmatrix}0.69&0.22\\-0.31&0.60\\-0.76&0.32\end{bmatrix}$, and $\mathbf{e}=\begin{bmatrix}P A D\end{bmatrix}^{\intercal}$.
 
\begin{figure}
\begin{floatrow}
\ffigbox{%
  \includegraphics[scale=0.20]{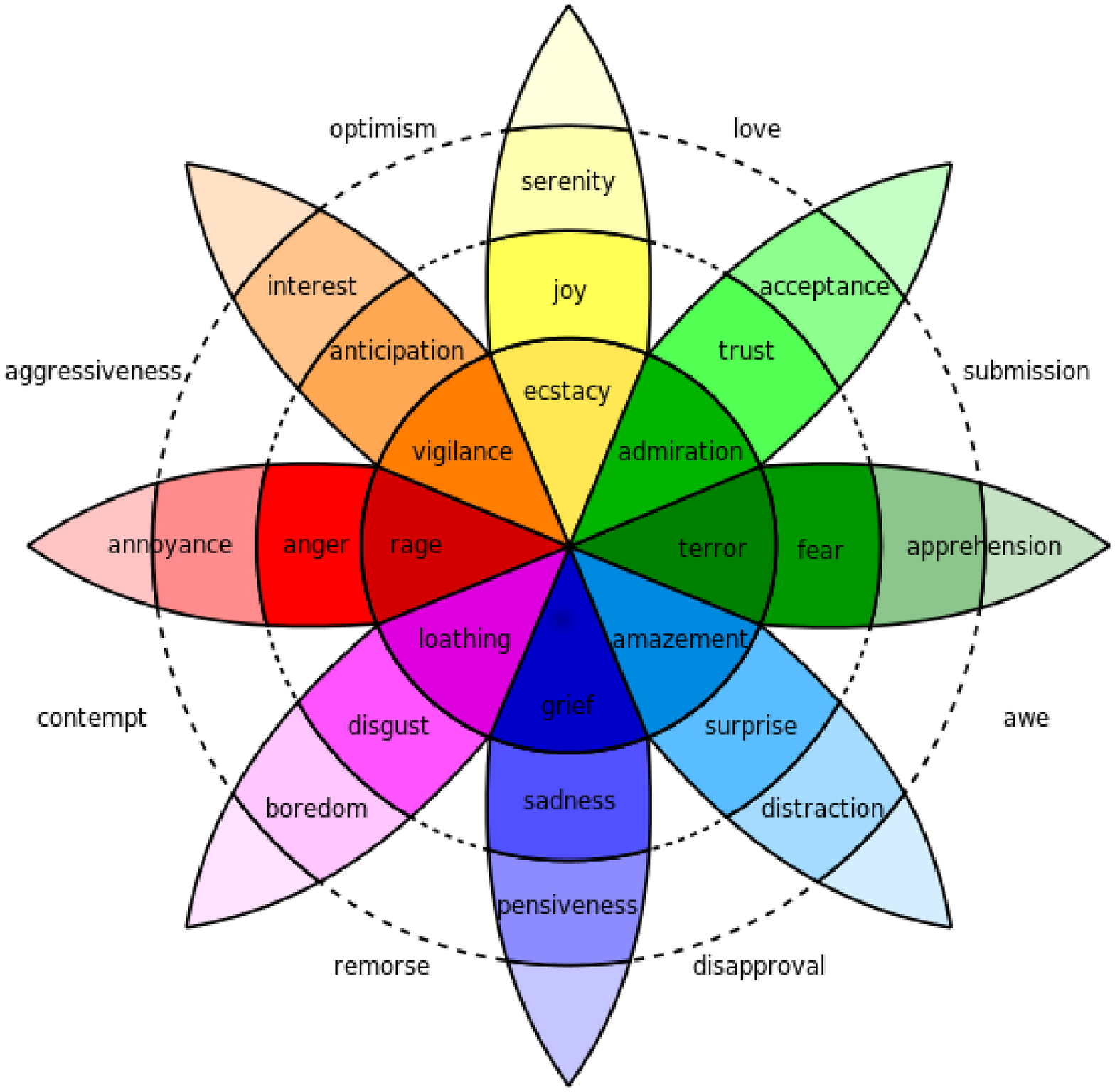}
}{%
  \caption{Plutchik's wheel. Relationships between emotions and colors. Hue is associated to a specific emotion, while saturation and luminance determine its intensity.}%
  \label{Fig:PlutchikWheel}
}
\capbtabbox{%
\scriptsize
\begin{tabular}{|c|c|c|c||c|c|c|}
\hline
\textbf{Emotion} & \textbf{H} & \textbf{S} & \textbf{L} & \textbf{P} & \textbf{A} & \textbf{D} \\
\hline
joy	    & 60	& 67    & 100 & 0.81 & 0.51 & 0.46 \\
ecstasy	            & 60	& 67	& 100 & 0.62 & 0.75 & 0.38 \\  
fear 	    & 120	& 100	& 59  & -0.64 & 0.60 & -0.43\\
terror	            & 120	& 100	& 50  & -0.62 & 0.82 & -0.43\\
amazement	        & 203	& 100	& 88  & 0.16 & 0.88 & -0.15\\
sadness 	& 240	& 68	& 100 & -0.63 & -0.27 & -0.33\\
boredom	            & 300	& 22	& 100 & -0.65 & -0.62 & -0.33\\
annoyance	        & 0	    & 45	& 100 & -0.58 & 0.40 & 0.01\\
anger 	    & 0	    & 100	& 100 & -0.51 & 0.59 & 0.25\\
interest	            & 29	& 45	& 100 & 0.64 & 0.51 & 0.17\\
vigilance	            & 29	& 100	& 100 & 0.49 & 0.57 & 0.45\\
\hline
\end{tabular}
}{%
  \caption{Color values of emotions according to the Plutchik's wheel and their associations to Mehrabian \cite{russell1977evidence} scores of Pleasure, Arousal and Dominance.}%
\label{Tab:PlutchikWheel}
 }
\end{floatrow}
\end{figure}

Eq.\ref{eq:part} provides a partial color mapping $\mathbf{E} \mapsto (S,L)$. To complete the picture we need to take into account hue values $H$. Unfortunately, the hue / PAD relation proposed in \cite{mehrabian1994} cannot be inverted.  We thus derive this component from Plutchik's psycho-evolutionary emotion theory \cite{Plutchik1980}. In his work, each emotion is associated to a given hue value, while saturation and luminance vary according to the emotion intensity (Fig. \ref{Fig:PlutchikWheel}). As we need an association between PAD and hue values, we rely on the classification made by Mehrabian \cite{russell1977evidence}, adopting the PAD values of a subset of corresponding affective states, as tabulated in Tab. \ref{Tab:PlutchikWheel}.    

Eventually, PAD values and the corresponding HSL values, can serve, respectively, as feature and target sets for learning the multivariate linear regression model given in Eq. \ref{eq:hslreg}. As in the case of Eq. \ref{eq:padreg} this is accomplished via ridge regression.

\begin{figure}[htb]
\centering
\includegraphics[scale=0.30]{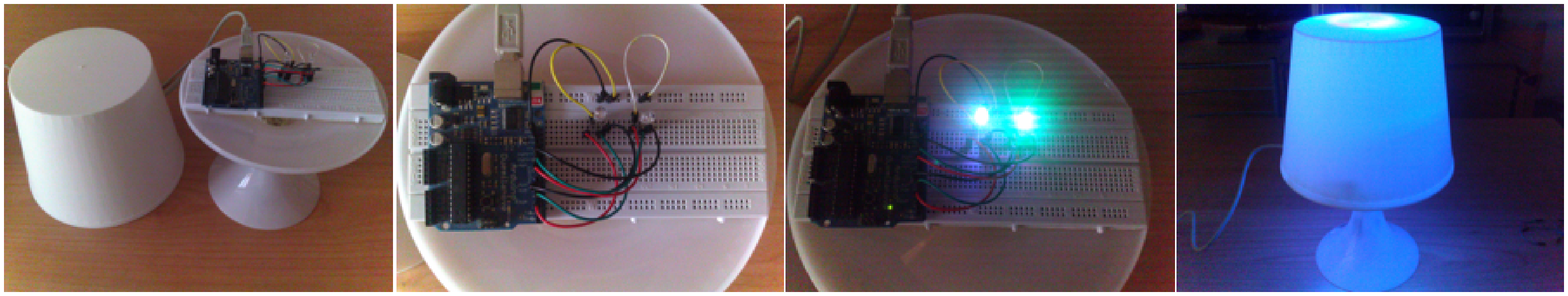}
\caption{Color control: the actual color stimulus is  generated through a modified  Ikea  lamp, where the RGB LED is controlled by an Arduino UNO board.} 
\label{Fig:FigArduino}
\end{figure}

Finally, the obtained HSL values are converted into RGB space. The latter step has a practical motivation.  As discussed from the beginning, the realisation of the  transcoding process has been experimented through the Mood Lamp,  an affective object conceived as i) a sensing interface, namely  a low-cost web camera / notebook  communicating via USB  with ii) a modified Ikea lamp, equipped with an Arduino UNO board to control an RGB LED (see Fig. \ref{Fig:FigArduino}).
  

\begin{figure}[htb]
\centering
\includegraphics[scale=0.20]{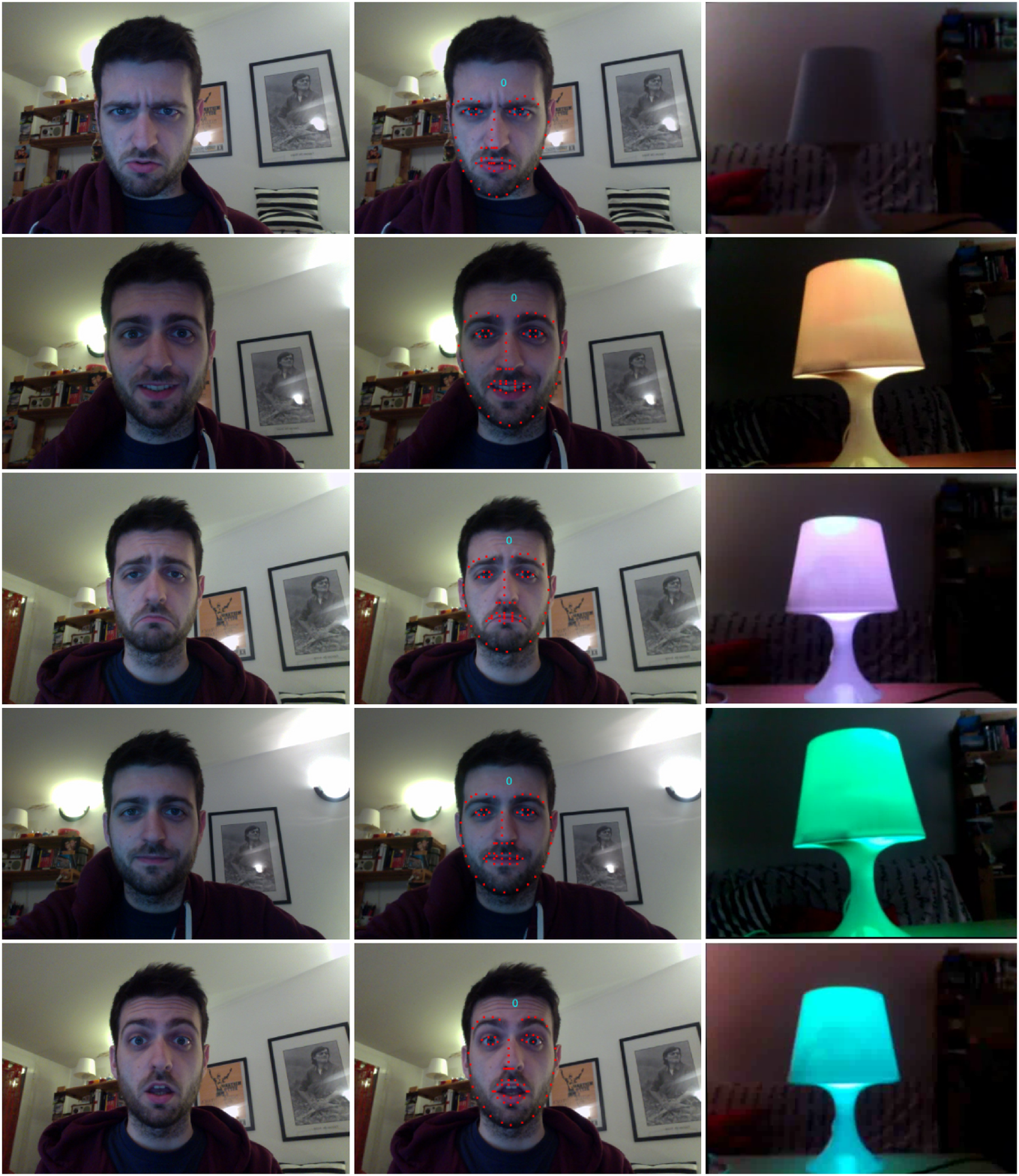}
\caption{Experimental results of transconding using the Mood Lamp.
} 
\label{Fig:MoodLTests}
\end{figure}




\section{Conclusion and further outlooks}
We have discussed how the IB framework provides a parsimonious and principled account of using a latent affect space to mediate the rechanneling between different stimuli.
As a final comment, we think it appropriate to remark that the general formalism  which is expounded here admits a far wider range of applicability than that to which it has been presented in this work. The framework could be usefully adopted for  current affective computing systems that more and more relying on the availability of different sensors (e.g., for monitoring autonomic activity) and brain interfaces \cite{Vinciarelli2013}.  Indeed, such systems are confronted with the issue  of  
finding relations in high-dimensional and heterogeneous data spaces,  one example being  data fusion 
among several others which would emerge from the application of this approach to concrete instances. 




\section*{Acknowledgments}
The  research  was  carried  out  as  part of the project "Interpreting emotions: a computational tool integrating facial expressions and biosignals based shape analysis and bayesian networks", supported by the Italian Government, managed by MIUR, nanced by the \emph{Future in Research} Fund.

\bibliographystyle{splncs03}
\bibliography{bibVittorio}

\end{document}